\documentclass[letterpaper]{article} % DO NOT CHANGE THIS
\usepackage{aaai25}  % DO NOT CHANGE THIS
\usepackage{times}  % DO NOT CHANGE THIS
\usepackage{helvet}  % DO NOT CHANGE THIS
\usepackage{courier}  % DO NOT CHANGE THIS
\usepackage[hyphens]{url}  % DO NOT CHANGE THIS
\usepackage{graphicx} % DO NOT CHANGE THIS
\usepackage{natbib}  % DO NOT CHANGE THIS AND DO NOT ADD ANY OPTIONS TO IT
\usepackage{caption} % DO NOT CHANGE THIS AND DO NOT ADD ANY OPTIONS TO IT
\usepackage{algorithm}
\usepackage{algorithmic}
\usepackage{enumitem}
\usepackage{amsmath, amsfonts, amssymb}  % For mathbb
\usepackage{graphicx}                    % For includegraphics
\usepackage{booktabs}
\usepackage{graphicx}
\usepackage{caption}
\usepackage{xcolor}
\usepackage{pifont} % 用于 √ 和 × 符号
\usepackage{newfloat}
\usepackage{listings}
\usepackage{colortbl}

\definecolor{mygray}{gray}{.9}

\DeclareCaptionStyle{ruled}{labelfont=normalfont,labelsep=colon,strut=off} % DO NOT CHANGE THIS
\floatstyle{ruled}
\newfloat{listing}{tb}{lst}{}
\floatname{listing}{Listing}
\title{IMAGDressing-v1: Customizable Virtual Dressing}
\author {
    Fei Shen\textsuperscript{\rm 1 },
    Xin Jiang\textsuperscript{\rm 1 },
    Xin He\textsuperscript{\rm 2},
    Hu Ye\textsuperscript{\rm 3},
    Cong Wang\textsuperscript{\rm 4},
    Xiaoyu Du\textsuperscript{\rm 1},
    Zechao Li\textsuperscript{\rm 1},
    Jinhui Tang\textsuperscript{\rm 1 \#}
}
\affiliations {
    \textsuperscript{\rm 1}Nanjing University of Science and Technology\\
    \textsuperscript{\rm 2}Wuhan University of Technology\\
    \textsuperscript{\rm 3}Tencent AI Lab\\
    \textsuperscript{\rm 4}Nanjing University\\
\textcolor{magenta}{\url{https://imagdressing.github.io/}}
}

\usepackage{bibentry}
\begin{document}

\maketitle

\begin{abstract}
Latest advances have achieved realistic virtual try-on (VTON) through localized garment inpainting using latent diffusion models, significantly enhancing consumers' online shopping experience. However, existing VTON technologies neglect the need for merchants to showcase garments comprehensively, including flexible control over garments, optional faces, poses, and scenes. To address this issue, we define a virtual dressing (VD) task focused on generating freely editable human images with fixed garments and optional conditions. Meanwhile, we design a comprehensive affinity metric index (CAMI) to evaluate the consistency between generated images and reference garments.
Then, we propose IMAGDressing-v1, which incorporates a garment UNet that captures semantic features from CLIP and texture features from VAE. 
We present a hybrid attention module, including a frozen self-attention and a trainable cross-attention, to integrate garment features from the garment UNet into a frozen denoising UNet, ensuring users can control different scenes through text. 
IMAGDressing-v1 can be combined with other extension plugins, such as ControlNet and IP-Adapter, to enhance the diversity and controllability of generated images. 
Furthermore, to address the lack of data, we release the interactive garment pairing (IGPair) dataset, containing over 300,000 pairs of clothing and dressed images, and establish a standard pipeline for data assembly.
Extensive experiments demonstrate that our IMAGDressing-v1 achieves state-of-the-art human image synthesis performance under various controlled conditions.
The code and model will be available at \textcolor{magenta}{\url{https://github.com/muzishen/IMAGDressing.}}
\end{abstract}


\section{Introduction}
Virtual dressing (VD) aims to achieve comprehensive and personalized clothing displays for merchants by utilizing given garments and optional faces, poses, and descriptive texts. This technology holds significant potential for practical applications in e-commerce and entertainment.
However, existing works primarily focus on virtual try-on (VTON)~\cite{han2018viton, choi2021viton, kim2024stableviton, morelli2023ladi, xu2024ootdiffusion, yang2024texture} tasks for consumers, which involve given garments and fixed human conditions, lacking flexibility and editability. While VD offers greater freedom and appeal, it also presents more significant challenges.

\begin{figure}[t]
%\vspace{-0.2cm}
  \centering
  \includegraphics[width=\linewidth]{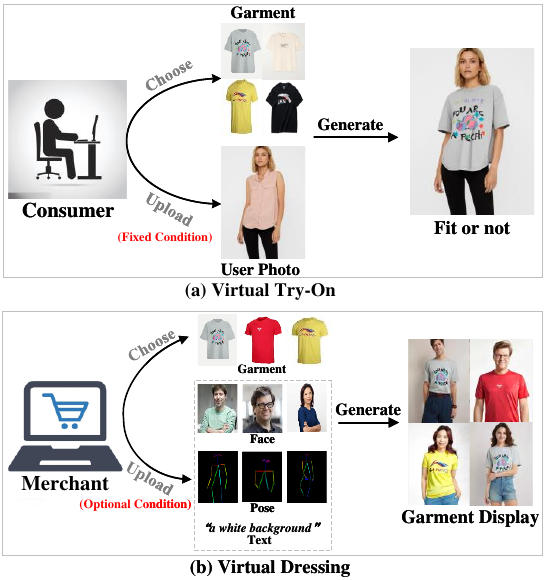}
  \caption{Differences between virtual try-on  and virtual dressing tasks in conditions and applicable scenarios.
  }
  \label{fig:example}
  \vspace{-10pt}
\end{figure}

To enhance the shopping experience for consumers in e-commerce, VTON~\cite{han2018viton, choi2021viton} tasks have become increasingly popular within the community. Early methods primarily relied on generative adversarial network (GAN)~\cite{creswell2018generative}, typically comprising a warping module to learn the semantic correspondence between clothes and the human body, and a generator module to synthesize the warped clothes onto the person's image.
However, GAN-based methods~\cite{choi2021viton, han2018viton} often suffer from instability due to the min-max training objective, and they have limitations in preserving texture details and handling complex backgrounds. 
Recently, latent diffusion models~\cite{ramesh2022hierarchical, zhang2023adding, saharia2022photorealistic} have made significant advances in VTON applications. These methods~\cite{kim2024stableviton, zeng2024cat, xu2024ootdiffusion} better retain the texture information of the input garments through a multi-step denoising process, ultimately generating images of specific individuals wearing the target clothing.
Nevertheless, as illustrated in Figure \ref{fig:example} (a), VTON is essentially a local image inpainting task for consumer scenarios, requiring only the faithful preservation of given garment features. 
It overlooks the need for comprehensive clothing displays in merchant scenarios, lacking the ability to customize faces, poses, and scenes.

To address this, as illustrated in Figure \ref{fig:example} (b), we define a virtual dressing (VD) task aimed at generating freely editable human images with fixed garment and optional conditions, and then design a comprehensive affinity metric index (CAMI) to evaluate the consistency between generated images and reference garments.
\underline{The differences between VTON and VD are as follows}:
(1) \textit{\textbf{User Experience.}} VTON synthesizes images based on given clothing and specific person conditions, providing users with a static experience of partial inpainting. 
In contrast, VD centers on clothing and combines it with optional conditions to synthesize images, offering users a more interactive and flexible experience.
(2) \textit{\textbf{Application Scenarios.}} VTON is primarily used for personalized services for consumers, such as trying on clothes online to see if they suit them. 
In comparison, VD is mainly used by merchants on e-commerce platforms to showcase clothing, providing a comprehensive view of clothing ensembles.
(3) \textit{\textbf{Accuracy Requirements.}} VTON focuses on ensuring natural transitions and detailed handling between the clothing and the given model's body. 
Building on these requirements, VD further emphasizes the uniformity and aesthetics of clothing displays under given clothing conditions and optional elements.

Furthermore, this paper presents IMAGDressing-v1, a latent diffusion model specifically designed for custom virtual dressing for merchants. 
IMAGDressing-v1 consists primarily of a trainable garment UNet and a denoising UNet. 
Since the VAE can nearly losslessly reconstruct images, the garment UNet is used to simultaneously capture semantic features from CLIP and texture features from the VAE.
The denoising UNet introduces a hybrid attention module, comprising a frozen self-attention and a trainable cross-attention modules, to integrate clothing features from the clothing UNet into it. This integration allows users to control different scenes through text prompts.
Moreover, IMAGDressing-v1 can be combined with other extensions, such as ControlNet~\cite{zhang2023adding} and IP-Adapter~\cite{ipadapter}, to enhance the diversity and controllability of generated images. Lastly, to address the issue of data scarcity, we have collected and released the large-scale interactive garment pairing (IGPair) dataset, containing over 300,000 pairs of clothing and dressed images. 
The contributions of this paper are summarized as follows:
\begin{itemize}[label=\textbullet]
    \item This paper introduces a new virtual dressing (VD) task for merchants and designs a comprehensive affinity measurement index (CAMI) to evaluate the consistency between generated images and reference garment.
    \item We propose IMAGDressing-v1, which includes a garment UNet for extracting fine-grained clothing features and a denoising UNet with a hybrid attention module to balance clothing features with text prompt control.
    \item IMAGDressing-v1 can be combined with other extensions, such as ControlNet and IP-Adapter, to enhance the diversity and controllability of generated images.
    \item We collect and release a large-scale interactive garment pairing (IGPair) dataset, containing over 300,000 pairs, available for the community to explore and research.

\end{itemize}

\section{Related Work}
\subsection{Virtual Try-On}
Early virtual try-on~\cite{dong2020fashion, jo2019sc, lee2020maskgan, liu2020mgcm, choi2021viton} typically utilized generative adversarial networks (GANs)~\cite{creswell2018generative} and a two-stage strategy. 
Initially, they would warp the clothing to the desired shape, then use a GAN-based generator to merge the warped clothing onto the human model. 
For instance, VITON-HD~\cite{choi2021viton} addresses issues of clothing-body occlusion and mismatch by performing warping and segmentation simultaneously. 
GP-VTON~\cite{xie2023gp} introduces local warping and global parsing to independently model the deformation of different clothing regions, aiming for a more accurate fit. 
To achieve precise clothing deformation, some methods~\cite{han2019clothflow,lee2022high} estimate a dense flow map to guide the reshaping process. 
Additionally, some approaches~\cite{ge2021parser, issenhuth2020not} use normalization or distillation strategies to address the misalignment between the warped clothing and the human body. 
However, GAN-based methods face instability due to the min-max nature of their training objectives and have limitations in preserving texture details and handling complex backgrounds.

Recent research~\cite{morelli2023ladi, gou2023taming, zhu2023tryondiffusion} have incorporated pre-trained diffusion models as priors for VTON tasks. 
For example, LADI-VTON~\cite{morelli2023ladi} and DCI-VTON~\cite{gou2023taming} explicitly warp clothes to align them with the human body, then use diffusion models to merge the clothes with the body. TryOnDiffusion~\cite{zhu2023tryondiffusion} proposed an architecture with two parallel UNets and demonstrated the capability of diffusion-based virtual try-on by training on large-scale datasets. Similarly, OOTDiffusion~\cite{xu2024ootdiffusion} and IDM~\cite{choi2024improving} utilize parallel UNets for garment feature extraction and enhance integration through self-attention. 
StableVITON~\cite{kim2024stableviton} introduces a zero-initialized cross-attention block to inject intermediate features of the spatial encoder into the UNet decoder.
While diffusion-based VTON methods can combine clothing with a fixed model, producing fine-grained static images, VTON is essentially a local image inpainting task tailored for consumer scenarios, merely needing to faithfully preserve the given clothing features. As previously mentioned, VTON overlooks the need for comprehensive garment presentation in commercial contexts and cannot customize faces, poses, and scenes.

\subsection{Latent Diffusion Model}
While latent diffusion models (LDMs)~\cite{rombach2022high, wang2024ensembling, shen2024boosting, wang2024ensembling} have been widely used for text-to-image (T2I) generation and editing tasks, the inaccuracy of natural language limits fine-grained image control. Various methods have been proposed to add conditional control to T2I diffusion models to address this. 
For example, ControlNet~\cite{zhang2023adding} and T2I Adapter~\cite{mou2024t2i} introduce additional conditional encoding modules, such as edges, depth, and human poses, to control diffusion models and text prompts. 
IP-Adapter~\cite{ipadapter} conditions T2I diffusion models on high-level semantics of reference images, using both text and visual cues to guide image generation.
Uni-ControlNet~\cite{zhao2024uni} proposes a unified framework that flexibly and composably handles different conditional controls within a single model to reduce computational costs. MasaCtrl~\cite{cao2023masactrl} achieves consistent image generation and complex non-rigid image editing through self-attention transformation without additional training costs. Similarly, InstructPix2Pix~\cite{brooks2023instructpix2pix} retrains LDMs by adding extra input channels to the first convolutional layer to follow editing instructions. 
PCDMs~\cite{shen2023advancing} proposes proposes a multi-stage conditional diffusion model for pose guided character image generation.
In this paper, we leverage the capabilities of frozen LDMs in text-to-image generation to achieve garment-centric image generation and editing.

\begin{table}[t]
    \centering
    \captionsetup{font=small}

    \setlength{\tabcolsep}{2pt} % 调整列间距
    \renewcommand{\arraystretch}{1.1} % 调整行高
    \small % 调整字体大小
    \begin{tabular}{lccccc}
        \toprule
        Dataset & Public & Caption  & \# Garments & \# Pairs & Resolution \\
        \midrule
        TryOnGAN &\ding{55} & \ding{55} & 52,000 & 52,000 & $512 \times 512$ \\
        Revery AI &\ding{55} & \ding{55} & 321,000& 321,000 & $512 \times 512$ \\
        VITON-HD  &\ding{51} & \ding{55} & 13,679 & 13,679 & $1024 \times 768$ \\
        Dress Code & \ding{51} & \ding{55} & 53,792 & 53,792 & $1024 \times 768$ \\
        \textbf{IGPair (Ours)}  & \ding{51} & \ding{51} &86,873  &324,857  & $> 2K\times2K$ \\
        \bottomrule
    \end{tabular}
        \vspace{-0.3cm}
        \caption{Comparison between IGPair and the widely used datasets.}\label{dataset}
\end{table}

\section{IGPair Dataset}
\ding{228} \textbf{Q1: \textit{What kind of data is suitable for VD task?}} ~We identify three critical requirements for an ideal virtual dressing dataset: (1) it should be publicly accessible for research purposes; (2) it should include high-resolution images of both garment and models wearing the clothing; (3) it should encompass a variety of scenes and styles, with detailed textual descriptions. 
As shown in Table~\ref{dataset}, the proposed IGPair dataset not only meets all the aforementioned requirements but also provides six times the number of image pairs compared to the largest publicly available dataset, VITON-HD~\cite{choi2021viton}, surpassing TryOnGAN~\cite{lewis2021tryongan}, Revery AI~\cite{li2021toward}, VITON-HD~\cite{choi2021viton}, and Dress Code~\cite{morelli2022dress} datasets.
Notably, IGPair includes multiple models for each clothing item. It is also the only dataset with a resolution exceeding $2K\times2K$. 
Additionally, IGPair is the only publicly available dataset that includes textual descriptions, diverse scenes, and various styles.

\begin{figure}[t]
    \centering
    \includegraphics[width=\linewidth]{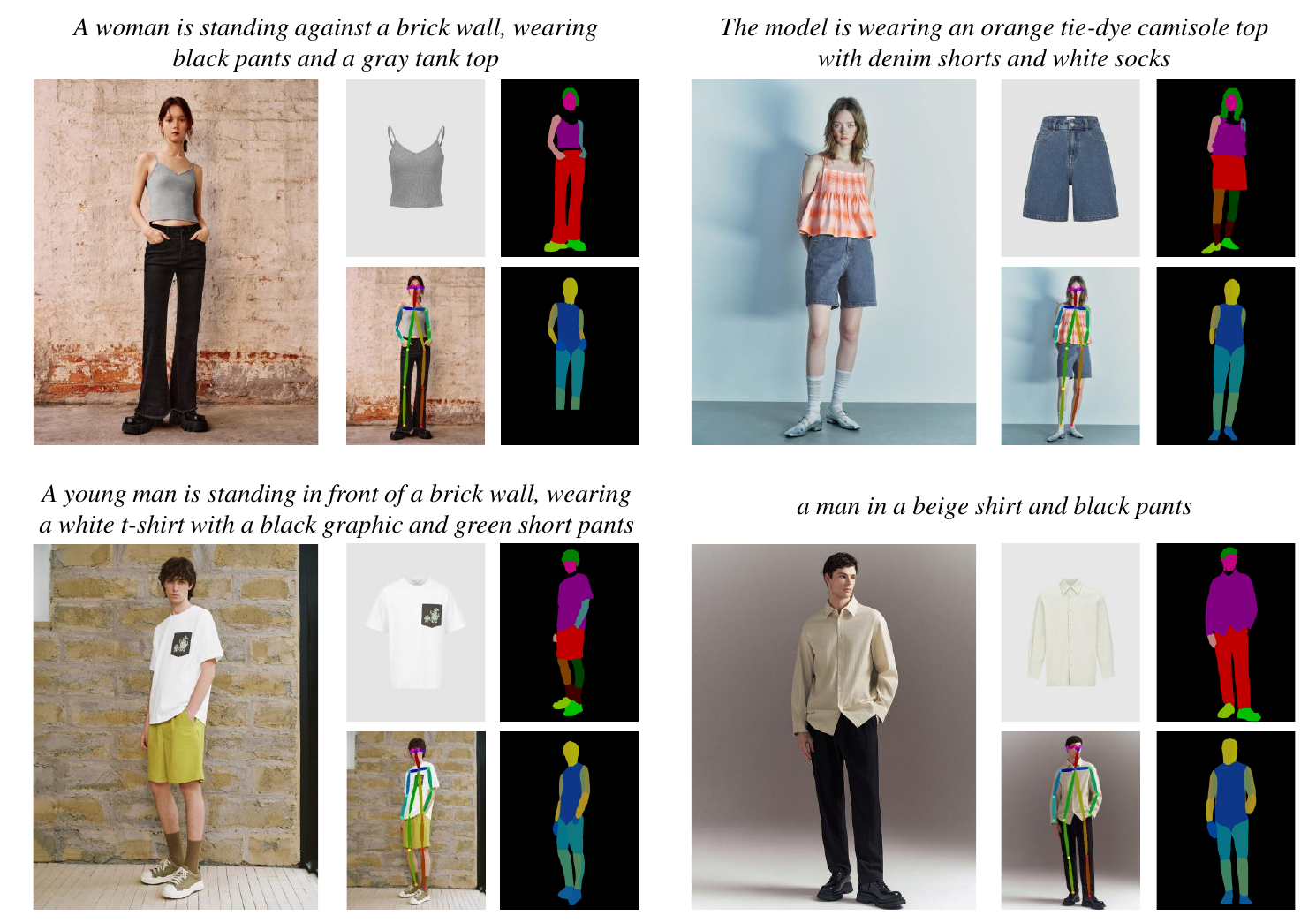}
    \caption{Sample pairs from the IGPair dataset, including pose keypoints, dense poses, and human body segmentation masks. More sample refer to supplementary material.}
      \label{fig:igpair}
      \vspace{-0.3cm}
\end{figure}

\noindent \ding{228} \textbf{Q2: \textit{How is the IGPair dataset collected and annotated?}} ~All images are sourced from the internet and encompass a variety of clothing styles, including casual, formal, athletic, fashionable, and vintage, etc. 
Initially, we collected 500,000 garment images, each accompanied by 2 to 5 images of people wearing the clothing from different perspectives.
We then use classifiers to differentiate between clothing and human and employ a human pose estimator to select complete and usable images of clothing on models. 
After this automated stage, we manually verified all images. 
We categorize the garment into 18 types, and the dataset consists of 324,857 image pairs.
To further enrich our dataset, we use OpenPose~\cite{cao2017realtime} to extract 18 key points for each human figure, DensePose~\cite{guler2018densepose} to compute dense poses for each reference model, and SCHP~\cite{li2020self} to generate segmentation masks for body parts and clothing items. 
We utilize BLIP2-OPT-6.7B~\cite{li2023blip}, INTERNLM-XCOMPOSER2-VL-7B~\cite{internlmxcomposer2}, LLaVA-V1.5-13B~\cite{liu2024improved}, and Qwen-VL-Chat~\cite{Qwen-VL} to generate captions for the images.
All model images are anonymized. 
Samples of human models and clothing pairs from our dataset, along with the corresponding additional information, are shown in Figure~\ref{fig:igpair}.
More detail refer to supplementary material.

\begin{figure*}[t]
%\vspace{-0.2cm}
  \centering
\includegraphics[width=1.0\linewidth]{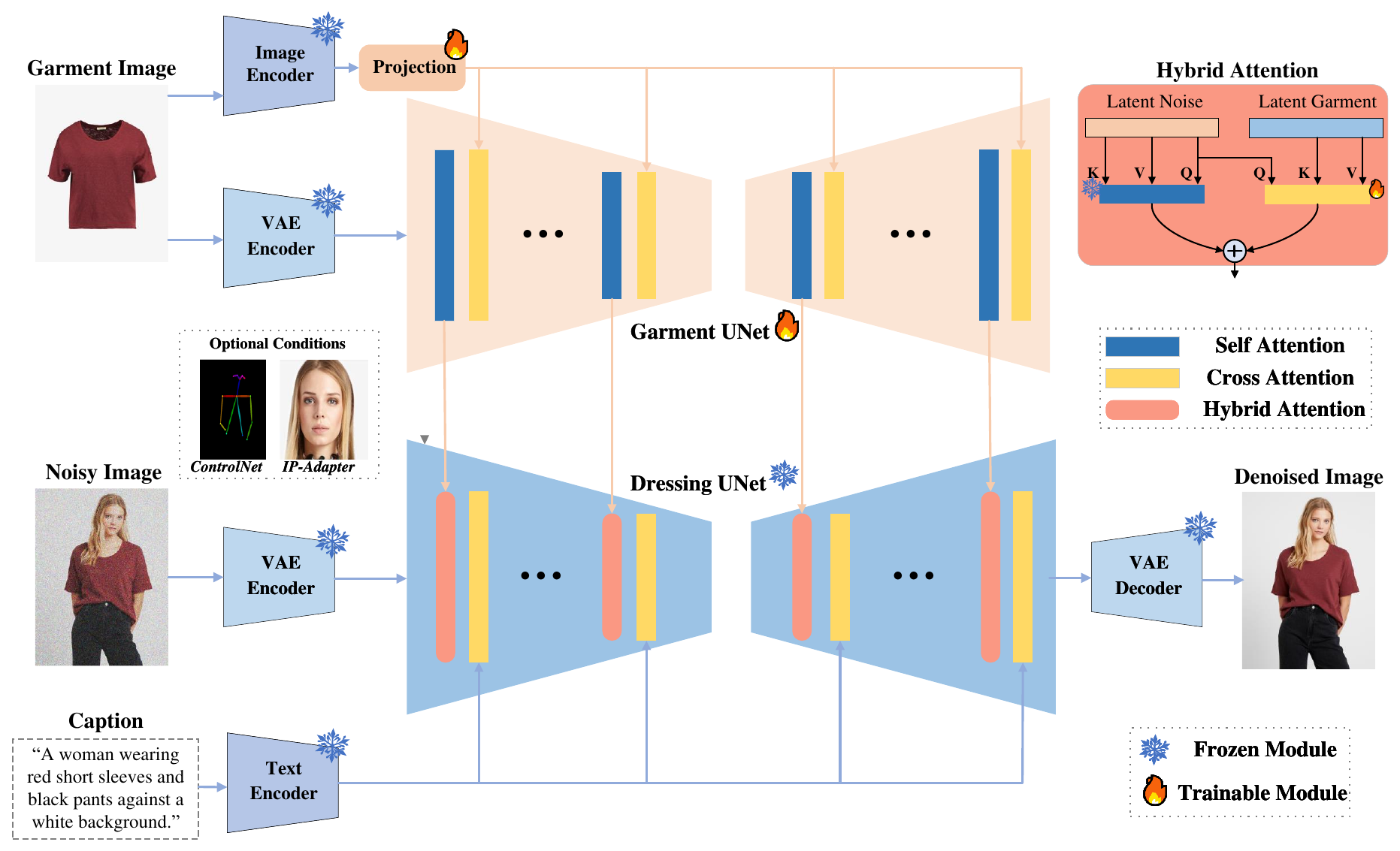}
  \caption{Illustration of the proposed IMAGDressing-v1 framework. It mainly consists of a trainable garment UNet and a frozen denoising UNet. The former extracts fine-grained garment features, while the latter balances these features with text prompts. IMAGDressing-v1 is compatible with other community modules, such as ControlNet and IP-Adapter.
  }
  \label{fig:pipeline}
  \vspace{-10pt}
\end{figure*}

\noindent \ding{228} \textbf{Q3: \textit{How to evaluate the consistency between the generated image and the garment?}} ~We propose a comprehensive affinity metric index (CAMI) for evaluating VD task, which includes the unspecified score (CAMI-U) and the specified score (CAMI-S). CAMI-U represents the score for image generation without specified pose, face, and text scenarios. In contrast, CAMI-S represents the score for image generation with the specified pose, face, and text scenarios. CAMI-U focuses on the clothing images' structure $S_s$, texture $S_t$, and keypoints $S_k$. 
CAMI-S builds upon CAMI-U by adding pose matching degree $S_p$, facial similarity $S_f$, and text-image matching degree $S_c$.
\vspace{-3pt}
\begin{align}
S_{\text{CAMI-U}} &= S_s + S_t + S_k, \\
S_{\text{CAMI-S}} &= S_{\text{CAMI-U}} + S_{p} + S_{f} + S_{c}.
\vspace{-3pt}
\end{align}
More detailed settings are to be provided in the supplementary material. 
Additionally, we also utilize MP-LPIPS~\cite{magiccloth}, and ImageReward~\cite{xu2024imagereward} to evaluate the quality of the generated images.

\section{Methodology}
% In this section, we first introduce the fundamental concepts of latent diffusion models. Next, we present our proposed IMAGDressing-v1 framework for customizable virtual dressing. Finally, we describe the large-scale interactive garment pairing (IGPair) dataset that we have collected.
\subsection{Preliminaries}\label{pre}
Unlike other pixel-based diffusion models, latent diffusion models (LDMs)~\cite{rombach2022high} aim to perform the denoising process in the latent space to reduce computational costs. 
An LDM typically comprises a variational autoencoder (VAE)~\cite{kingma2013auto}, a CLIP text encoder~\cite{radford2021learning}, and a denoising UNet. The VAE transforms images into latent space representations and vice versa. 
Specifically, the VAE encoder $\mathcal{E}$ compresses the original image $x$ into a latent representation $z$, i.e., $z = \mathcal{E}(x)$, while the VAE decoder $\mathcal{D}$ reconstructs the image $x$ from the latent representation $z$, i.e., $x = \mathcal{D}(z)$. The CLIP text encoder converts text prompts into token embeddings $c$. During the diffusion process, Gaussian noise ${\epsilon}$ is added to the latent representation z over timestep $t$ to produce $z_t$, where ${t\in [0, T]}$. 
The denoising UNet then iteratively denoises $z_t$ during the denoising process.
To learn such a denoising UNet ${\epsilon}_\theta$ parameterized by ${\theta}$, for each timestep ${t}$, the training objective usually adopts a mean square error loss ${L_{LDM}}$, as follows,
% \vspace{-3pt}
\begin{equation}
L_{LDM} = \mathbb{E}_{\mathbf{z}_t, \epsilon \sim \mathcal{N}(\mathbf{0}, \mathbf{I}), \mathbf{c}, t}\left\|\epsilon_\theta\left(\mathbf{z}_t, \mathbf{c}, t\right)-\epsilon_t\right\|^2,
% \vspace{-3pt}
\end{equation}
where $\mathbf{z}_t=\sqrt{\alpha_t} \mathbf{z}_0+\sqrt{1-\alpha_t} \epsilon_t$ is the noisy latent at timestep $t$ and $\epsilon_t$ is the added noise. 
Here, $\mathbf{z}_0=\mathcal{E}\left(\mathbf{x}_0\right)$ and ${x}_{0}$ represents the real data with a text condition ${c}$.

During the sampling stage, the predicted noise is calculated using both the conditional model ${\epsilon}_{\theta}({x}_t, {c}, t)$ and the unconditional model ${\epsilon}_{\theta}({x}_t, t)$ via classifier-free guidance~\cite{ho2022classifier}.
% \vspace{-5pt}
\begin{equation}
\hat{{\epsilon}}_{\theta}({x}_t, {c}, t) = w {\epsilon}_{\theta}({x}_t, {c}, t) + (1 - w) {\epsilon}_{\theta}({x}_t, t).
% \vspace{-5pt}
\end{equation}
Here, ${w}$ is the guidance scale used to adjust the influence of the condition ${c}$.

\subsection{IMAGDressing-v1}\label{imag}
As shown in Figure~\ref{fig:pipeline}, the proposed IMAGDressing-v1 mainly consists of a trainable garment UNet, architecturally same Stable Diffusion V1.5 (SD v1.5)~\footnote{https://huggingface.co/runwayml/stable-diffusion-v1-5}. 
The difference lies in the garment UNet's ability to simultaneously capture garment semantic features from CLIP and texture features from VAE, since the VAE can nearly losslessly reconstruct images.
The lower part is a frozen denoising UNet, similar to SD v1.5, used for denoising the latent image under conditions. Unlike SD v1.5, we replace all self-attention modules with hybrid attention modules to more easily integrate garment features from the garment UNet and leverage the existing text-to-image capabilities for scene control via text prompts. 
Additionally, IMAGDressing-v1 includes an image encoder and projection layer for encoding garment features, as well as a text encoder for encoding textual features.
\subsubsection{Garment UNet.} Extracting fine-grained garment features is crucial for maintaining the consistency of garment details in VD task. 
To achieve this, the proposed garment UNet simultaneously extracts semantic information and texture features as garment characteristics. Specifically, given a garment image $\mathcal{X} \in \mathbb{R}^{3 \times H \times W}$, we first convert it into a latent space representation $\mathbf{Z}_g \in \mathbb{R}^{4 \times \frac{H}{8} \times \frac{W}{8}}$ using a frozen VAE Encoder~\footnote{https://huggingface.co/stabilityai/sd-vae-ft-mse}. 
Simultaneously, token embeddings are extracted from $\mathcal{X}$ using a frozen CLIP image encoder~\footnote{https://huggingface.co/laion/CLIP-ViT-H-14-laion2B-s32B-b79K} and a trainable projection layer, where we utilize a Q-Former~\cite{li2023blip} as the projection layer. 
Subsequently, the garment features from garment UNet interact thoroughly in the cross-attention mechanism, similar to the interaction between text and image in the original T2I model. 
Finally, the garment UNet aligns parallelly with the denoising UNet, injecting fine-grained features into the denoising UNet through hybrid attention. 
It is important to note that the garment UNet is only used to encode the reference image. Therefore, no noise is added to the reference image, and only a single forward pass is performed during the diffusion process.

\subsubsection{Hybrid Attention.} For VD task, an ideal denoising UNet should possess two key capabilities: (1) maintaining the original editing and generation abilities, and (2) incorporating additional garment features. The former can be achieved by freezing the modules of the denoising UNet, while the latter is accomplished through the proposed hybrid attention modules. Consequently, the architecture of the denoising UNet in IMAGDressing-v1 is similar to that of the original text-to-image SD v1.5 model, with the main difference being that we replace all self-attention modules in the denoising UNet with hybrid attention modules. 
As shown in Figure~\ref{fig:pipeline}, the hybrid attention module consists of a frozen self-attention module and a learnable cross-attention module. 
Here, the weights of the self-attention of hybrid attention module are initialized using the self-attention's weights from SD v1.5.
Assuming $\mathbf{Z}_d$ and $\mathbf{C}_g$ represent the query features and the garment features output by the garment UNet at corresponding positions, the output of hybrid attention $\mathbf{Z}_h$ can be defined as follows:
\begin{equation}
\mathbf{Z}_{h} = 
\underbrace{\operatorname{Softmax}\left(\frac{\mathbf{Q} \mathbf{K}^{\top}}{\sqrt{d}}\right) \mathbf{V}}_{\text{Self Attention}} + 
\lambda\underbrace{\operatorname{Softmax}\left(\frac{\mathbf{Q}\left(\mathbf{K}^{\prime}\right)^{\top}}{\sqrt{d}}\right) {V}^{\prime}}_{\text{Cross Attention}} ~,
\label{hyb}
\end{equation}
where $\lambda \in \mathbb[0, 1.5]$ is a hyperparameter used to regulate the strength of garment conditions.
$\mathbf{Q} =\mathbf{Z}_d\mathbf{W}_q, \quad 
\mathbf{K} = \mathbf{Z}_d\mathbf{W}_k, \quad 
\mathbf{V} = \mathbf{Z}_d \mathbf{W}_v, \quad$
$\mathbf{K}' = \mathbf{C}_g \mathbf{W}'_k$, 
and $\mathbf{V}' = \mathbf{C}_g \mathbf{W}'_v$. 
Here, $\mathbf{W}_q, \mathbf{W}_k, \mathbf{W}_v, \mathbf{W'}_k$, and $\mathbf{W'}_v$ are the weight matrices of the trainable linear projection layers. Noted that we share a query matrix $\mathbf{Q}$ for self attention and cross attention.
In the hybrid attention module, the self-attention is frozen while the cross-attention is trainable. In other words, in Eq.\ref{hyb}, only $\mathbf{W'}_k$ and $\mathbf{W'}_v$ are learnable. 
This approach allows us to retain the generative capabilities of the original T2I model, such as scene generation.

\subsubsection{Training and Inference.}
During training stage, we keep the parameters of the basic modules in the denoising UNet unchanged and only optimize the remaining modules. Let $\mathbf{C}_t$ represent the text condition, then the loss function $L_{LDM}$ is as follows,
\begin{equation}
L_{LDM} = \mathbb{E}_{\mathbf{z}_t, \epsilon \sim \mathcal{N}(\mathbf{0}, \mathbf{I}), \mathbf{C}_t, \mathbf{C}_g, t}\left\|\epsilon_\theta\left(\mathbf{z}_t, \mathbf{C}_t, \mathbf{C}_g, t\right)-\epsilon_t\right\|^2,
% \vspace{-3pt}
\end{equation}
In the inference stage, we also use classifier-free guidance according to Eq.~\ref{infer}.
\begin{equation}\label{infer}
\hat{{\epsilon}}_{\theta}({x}_t, \mathbf{C}_t, \mathbf{C}_g, t) = w {\epsilon}_{\theta}({x}_t, {\mathbf{C}_t, \mathbf{C}_g}, t) + (1 - w) {\epsilon}_{\theta}({x}_t, t).
% \vspace{-5pt}
\end{equation}
\noindent \ding{228} \textbf{Q4: \textit{How support customized generation?}}
As shown in Figure~\ref{fig:pipeline}, the weights of the basic modules are frozen in the denoising UNet, making the garment UNet essentially an adapter module compatible with other community adapters for customized face and pose generation.
For instance, to generate images of people in a given outfit and consistent pose, IMAGDressing-v1 can be combined with ControlNet-Openpose. To generate specific individuals wearing specified clothing, IMAGDressing-v1 can be integrated with the IP-Adapter. Furthermore, if both pose and face need to be specified simultaneously, IMAGDressing-v1 can be used in conjunction with both ControlNet-Openpose and IP-Adapter. Additionally, for VTON task, IMAGDressing-v1 also can be combined with ControlNet-Inpaint.

\begin{table}[t]
    \centering
    \captionsetup{font=small}
    \setlength{\tabcolsep}{1.6pt} % 调整列间距
    \renewcommand{\arraystretch}{1.0} % 调整行高
    \scriptsize % 调整字体大小
    \begin{tabular}{l||cccc}
        \hline
           \rowcolor{mygray}\cellcolor{mygray} Method & ImageReward ($\uparrow$) & MP-LPIPS ($\downarrow$)  & CAMI-U  ($\uparrow$) & CAMI-S  ($\uparrow$)\\
        \hline
        Blip-Diffusion &-2.224 & 0.1824 & 1.051 & -  \\
       Versatile Diffusion &-2.055 & 0.4321 & 1.253 & - \\
        IP-Adapter   &-2.267 & 0.4093 & 1.381 & -  \\
        MagicClothing  &-0.164 & 0.1499 & 1.655 & 2.692  \\
        \hline
        \textbf{Ours}   &\textbf{-0.095} & \textbf{0.1466} & \textbf{1.753} & \textbf{2.719}   \\
        \hline
    \end{tabular}
    \vspace{-0.3cm}
    \caption{Quantitative comparison of the IMAGDressing-v1 with several state-of-the-art methods.}\label{quan}
    \vspace{-0.5cm}
    \label{compare_metric}
\end{table}

\begin{figure*}[t]
    \centering
\includegraphics[width=1.0\textwidth]{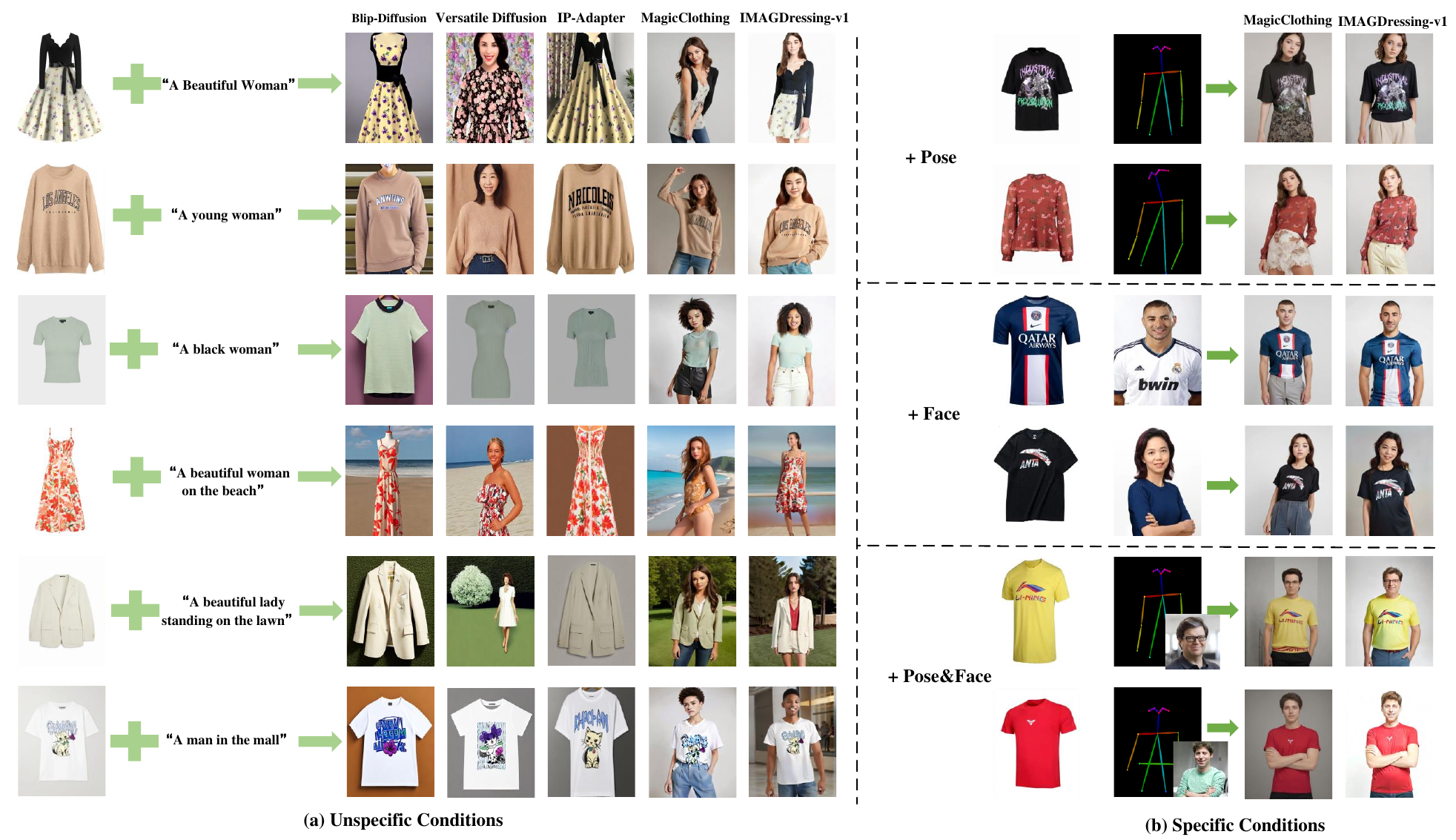}
    \vspace{-0.3cm}
    \caption{\small{Qualitative comparison with other SOTA methods under both unspecific and specific conditions, including BLIP-Diffusion~\cite{BLIP-Diffusion}, Versatile Diffusion~\cite{VersatileDiffusion}, IP-Adapter~\cite{ipadapter}, and MagicClothing~\cite{magiccloth}.}}
    \vspace{-0.5cm}
    \label{fig:compare_sota}
\end{figure*}

\section{Experiments} 
\subsection{Implementation Details}
In our experiments, we initialize the weights of our garment UNet by inheriting the pre-trained weights of the UNet in Stable Diffusion v1.5~\cite{sd15}, and finetune its weight.
Our model is trained on the paired images from the IGPair dataset at the resolution of $512~\times 640$.
We adopt the AdamW optimizer with a fixed learning rate of 5e-5. 
The model is trained for 200,000 steps on 10 NVIDIA RTX3090 GPUs with a batch size of 5.
At the inference stage, the images are generated with the UniPC sampler for 50 sampling steps and set the guidance scale $w$ to 7.0.
Please refer to the supplementary material for more details.
\subsection{Main Comparisons}
We compare our IMAGDressing-v1 with four state of-
the-art (SOTA) methods: Blip-Diffusion~\cite{BLIP-Diffusion}, Versatile Diffusion~\cite{VersatileDiffusion}, Versatile Diffusion~\cite{VersatileDiffusion}, and MagicClothing~\cite{magiccloth}. 

\begin{table}[t]
    \centering
    \captionsetup{font=small}
    \setlength{\tabcolsep}{1.2pt} % 调整列间距
    \renewcommand{\arraystretch}{1.1} % 调整行高
    \scriptsize % 调整字体大小
    \begin{tabular}{l||cccc}
        \hline
\rowcolor{mygray}\cellcolor{mygray}Method & ImageReward ($\uparrow$) & MP-LPIPS ($\downarrow$)  & CAMI-U ($\uparrow$) & CAMI-S ($\uparrow$)\\
        \hline
        A0 (Base) &-0.245 &0.1537 & 1.575 & 2.578  \\
       A1 (Base + IEB) &-0.178 & 0.1504 & 1.637 & 2.625 \\
         % w/o HA   &xx & xx & xx & xxx  \\
        \hline
        A2 (Base + IEB + HA)   &\textbf{-0.095} &\textbf{0.1466} & \textbf{1.753} & \textbf{2.719}   \\
        \hline
    \end{tabular}
    \vspace{-0.3cm}
    \caption{Quantitative results for different Settings. IEB and HA denote the image encoder branch and hybrid attention.} 
        \vspace{-0.3cm}
    \label{ablation_quan}
\end{table} 

\begin{figure}[t]
    \centering
\includegraphics[width=0.95\linewidth]{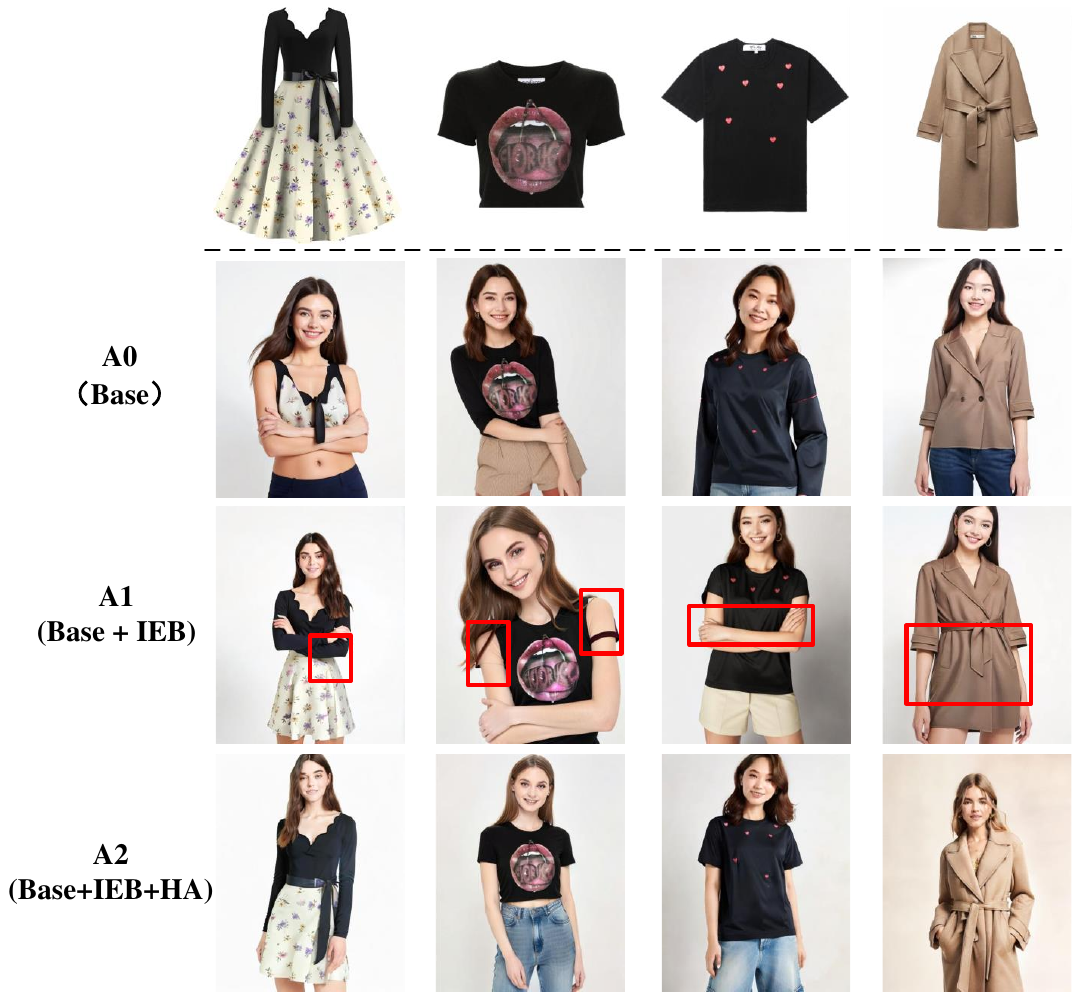}
\vspace{-0.3cm}
    \caption{Ablation study of each component.}
\vspace{-0.6cm}    
    \label{fig:ablation}
\end{figure}

\subsubsection{Quantitative Results.} 
As shown in Table~\ref{compare_metric}, since Blip-Diffusion~\cite{BLIP-Diffusion}, Versatile Diffusion~\cite{VersatileDiffusion}, and IP-Adapter~\cite{ipadapter} are not specifically designed VD models, they struggle to extract fine-grained garment features and generate character images that precisely match the text, pose, and garment attributes. 
This results in suboptimal performance across multiple metrics. Additionally, these models are incompatible with several plugins, making it impossible to compute the CAMI-S metric.
Compared to MagicClothing~\cite{magiccloth}, our IMAGDressing-v1 captures more detailed garment features through its image encoder branch and employs a hybrid attention mechanism. This mechanism integrates additional garment features while retaining the original text editing and generation capabilities. 
As a result, IMAGDressing-v1 demonstrates superior performance, outperforming other SOTA methods across all evaluation metrics.

\begin{figure*}[t]
    \centering
\includegraphics[width=\textwidth]{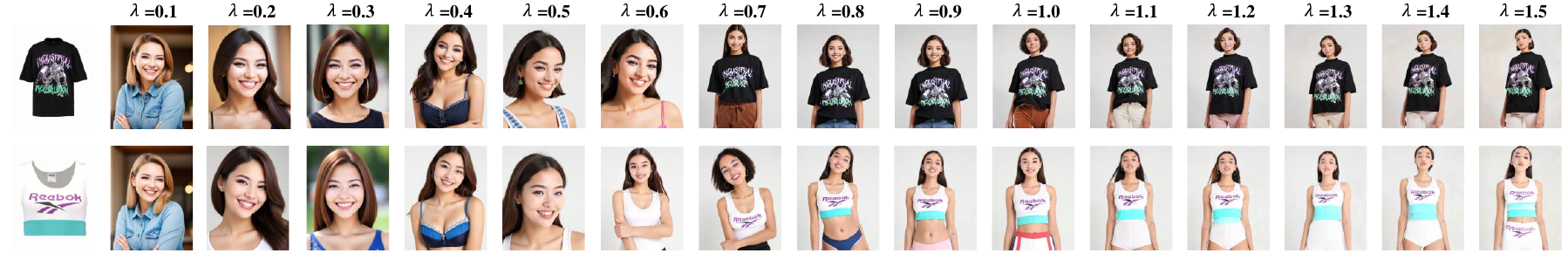}
% \vspace{-0.5cm}
    \caption{Example results with different garment strength $\lambda$.}
    % \vspace{-0.3cm}
    \label{fig:image_scale}
\end{figure*}

\begin{figure*}[t]
    \centering
\includegraphics[width=\textwidth]{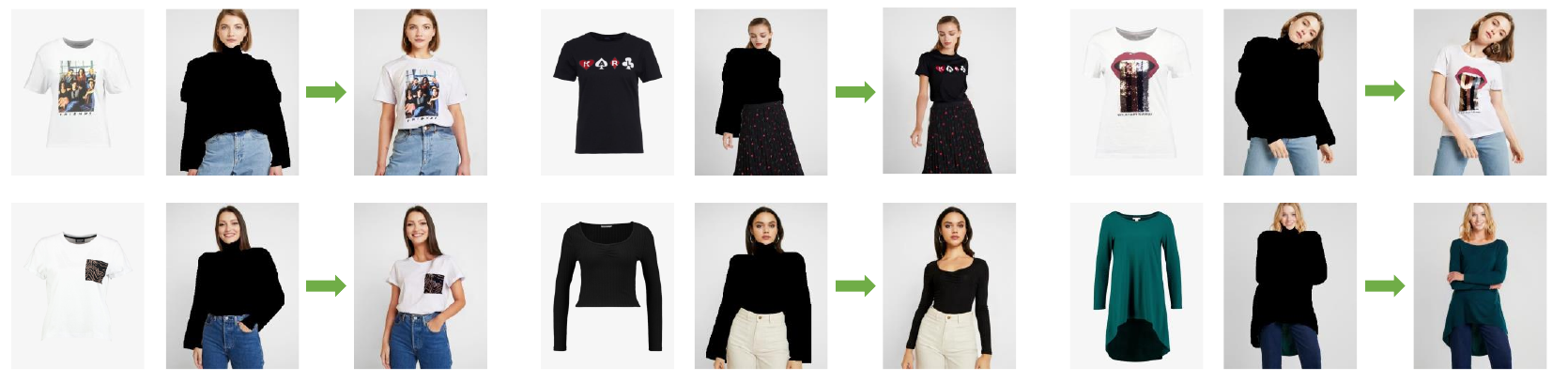}
\vspace{-0.5cm}
    \caption{Examples of plug-in results of our IMAGDressing-v1 combined with ControlNet-Inpaint for virtual try-on.}
    \vspace{-0.3cm}
    \label{fig:plug-ina}
\end{figure*}

\subsubsection{Qualitative Results.}
Figure~\ref{fig:compare_sota} illustrates the qualitative results of IMAGDressing-v1 compared to SOTA methods, including unspecific and specific condition generations.
In Figure~\ref{fig:compare_sota}(a), under unspecific conditions, BLIP-Diffusion~\cite{BLIP-Diffusion} and Versatile Diffusion~\cite{VersatileDiffusion} fail to faithfully reproduce garment textures. 
Although IP-Adapter maintains the overall appearance of the garments, it does not preserve the details well and, more importantly, does not follow the text prompts accurately. MagicClothing aligns closely with the text conditions; however, it struggles to retain the overall appearance and details of the garments, such as printed text or colors.
In contrast, IMAGDressing-v1 not only adheres to the text prompts but also preserves fine-grained garment details, demonstrating superior performance in VD tasks. Additionally, our method supports customized text prompt scenarios, as shown in the last three rows of Figure ~\ref{fig:compare_sota} (a).
Furthermore, Figure ~\ref{fig:compare_sota} (b) illustrates the qualitative results under specific conditions. We observe that IMAGDressing-v1 significantly outperforms MagicClothing in scenarios involving given poses, faces, or both. The results generated by IMAGDressing-v1 exhibit superior texture details and a more realistic appearance. This demonstrates the enhanced compatibility of IMAGDressing-v1 with community adapters, which enhances the generated images' diversity and controllability.

\subsection{Ablation Studies}
\subsubsection{Effectiveness of each component.} 
Table~\ref{ablation_quan} presents an ablation study to validate the effectiveness of the proposed image encoder branch (IEB) and hybrid attention (HA) module. Here, A0 (Base) denotes the setting without IEB and HA. We observe that A1, which uses IEB, shows improvements across all metrics, indicating that IEB effectively captures semantic garment features. Furthermore, A2 surpasses A1, demonstrating that the combination of IEB and HA further enhances quantitative results.
Additionally, Figure~\ref{fig:ablation} provides qualitative comparisons. We notice that A0 fails to adequately capture garment features in images with complex textures (2nd row). Although IEB (A1) partially addresses this issue, directly injecting IEB into the denoising UNet can lead to conflicts with the main model's features, resulting in obscured garment details~(3rd). Therefore, the HA module (A2) improves image fidelity by adjusting the intensity of garment details within the garment UNet (4th row), aligning with our quantitative results.
\subsubsection{Hyper-parameter $\lambda$.} In Figure~\ref{fig:image_scale} demonstrates the effects of the hyper-parameter $\lambda$ on generated samples with a fixed random seed. As $\lambda$ increases to 1.0, the garment in the generated character becomes more similar to the input garment. A smaller $\lambda$ ensures the generated results adhere more closely to the text prompts, while a larger $\lambda$ biases the results towards the input garment. This indicates that $\lambda$ effectively balances original editing and generation capabilities with additional garment features. Consequently, we empirically set $\lambda$ to 1.0 in our experiments.
\subsubsection{Potential application.}
Figure~\ref{fig:plug-ina} illustrates a potential application of IMAGDressing-v1 in virtual try-on (VTON). By combining IMAGDressing-v1 with ControlNet-Inpaint and masking the garment area, we achieve VTON. The results demonstrate that IMAGDressing-v1 can achieve high-fidelity VTON, showcasing significant potential.

\section{Conclusion}
While recent advancements in VTON using latent diffusion models have enhanced the online shopping experience, they fall short of allowing merchants to showcase garments comprehensively with flexible control over faces, poses, and scenes. To bridge this gap, we introduce the virtual dressing (VD) task, designed to generate editable human images with fixed garments under optional conditions. Our proposed IMAGDressing-v1 employs a garment UNet and a hybrid attention module to integrate garment features, enabling scene control through text. It supports plugins like ControlNet and IP-Adapter for greater diversity and controllability. Additionally, we release the IGPair dataset with over 300,000 pairs of clothing and dressed images, providing a robust data assembly pipeline. Extensive experiments validate that IMAGDressing-v1 achieves state-of-the-art performance in controlled human image synthesis.

\bibliography{aaai25}

\newpage

\appendix

\begin{figure*}[t]
    \centering
\includegraphics[width=\textwidth]{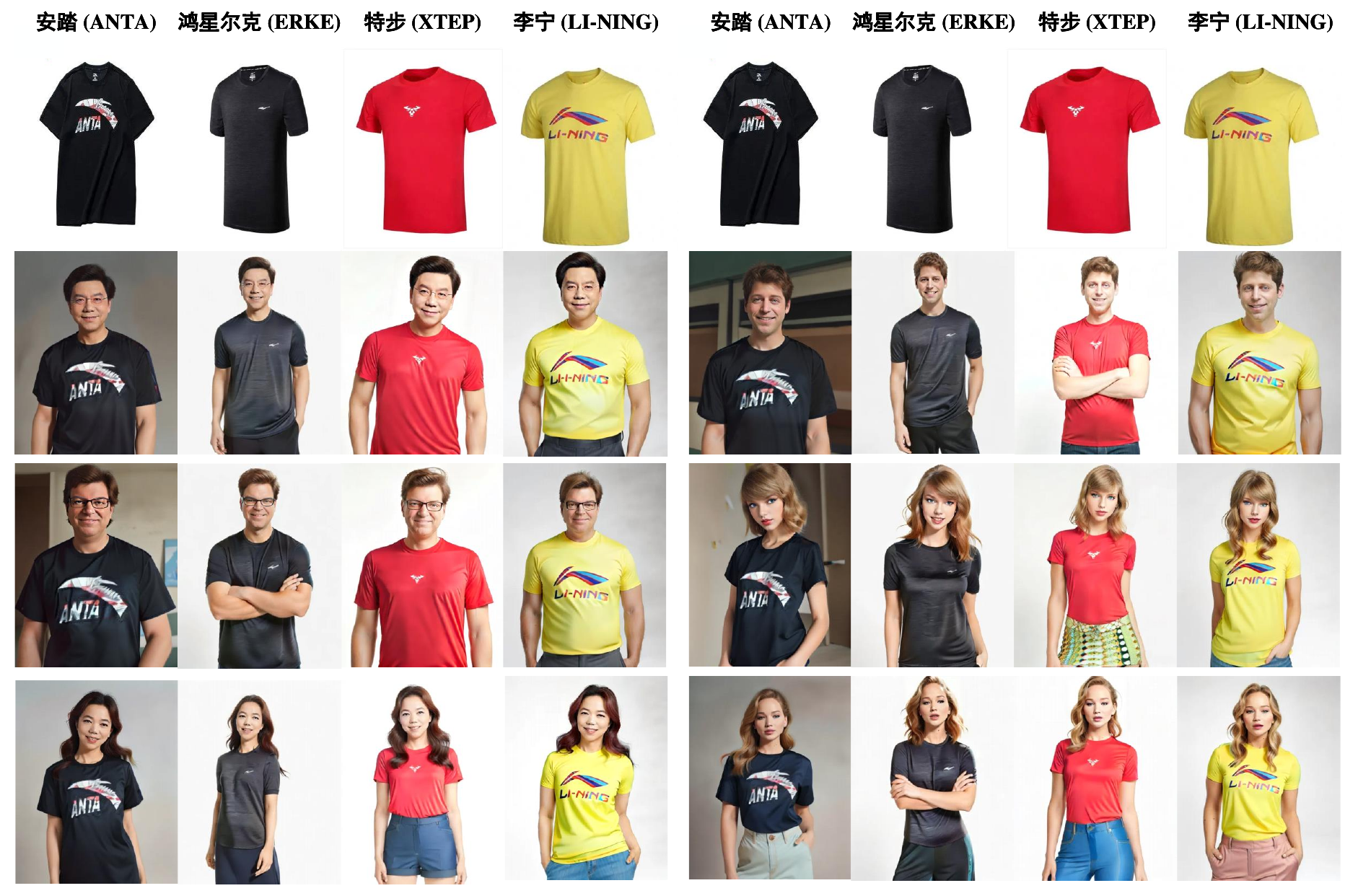}
\vspace{-0.5cm}
    \caption{More results of IMAGDressing-v1 synthesizing person images given garment with logos and optional faces.}
    \vspace{-0.3cm}
    \label{fig:more_face}
\end{figure*}
\begin{figure*}[t]
    \centering
\includegraphics[width=\textwidth]{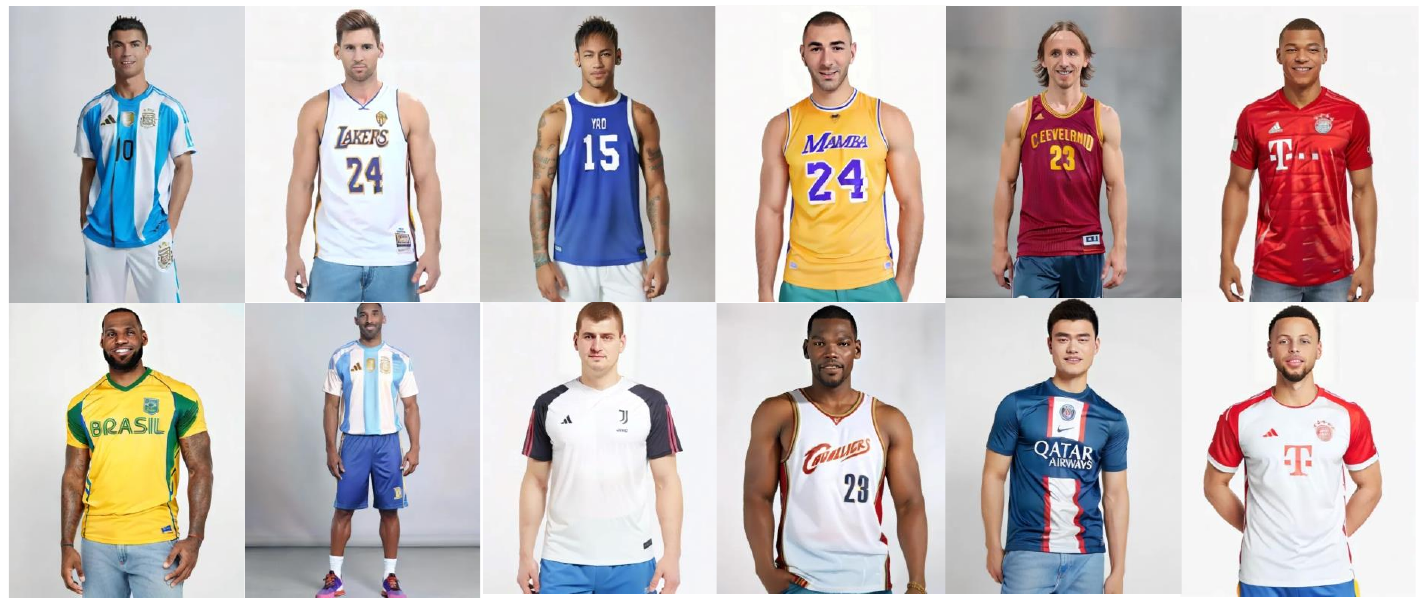}
\vspace{-0.5cm}
    \caption{More results of IMAGDressing-v1 generating person images with complex logos (such as text and dense letters) on garment and optional faces.}
    \vspace{-0.3cm}
    \label{fig:more_player_cloth}
\end{figure*}

\begin{figure*}[t]
    \centering
\includegraphics[width=0.95\textwidth]{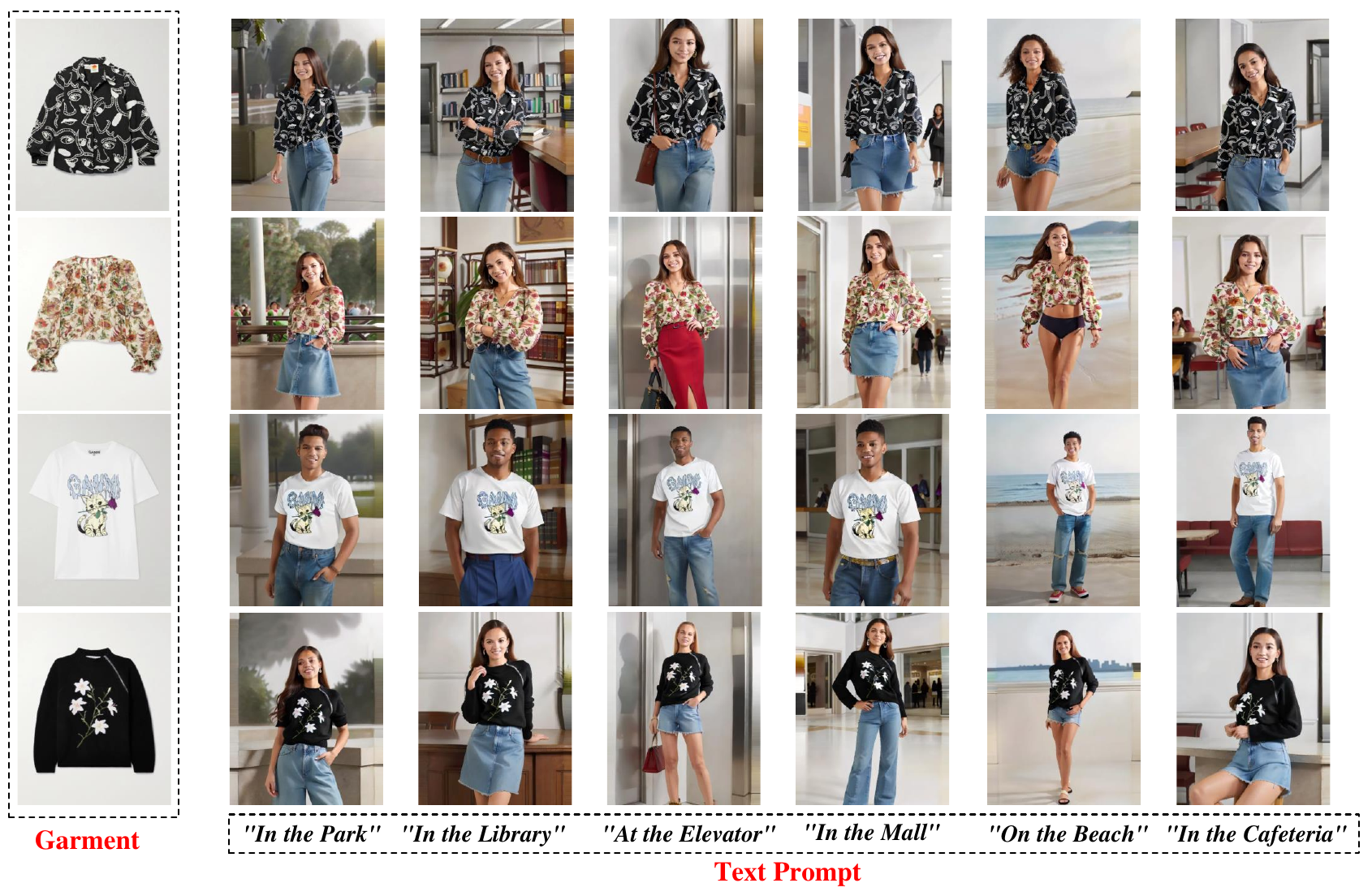}
\vspace{-0.3cm}
    \caption{More results of IMAGDressing-v1 synthesizing person images based on clothing and different text prompts.}
    \vspace{-0.3cm}
    \label{fig:more_text_promt}
\end{figure*}

\begin{figure*}[t]
    \centering
\includegraphics[width=0.95\textwidth]{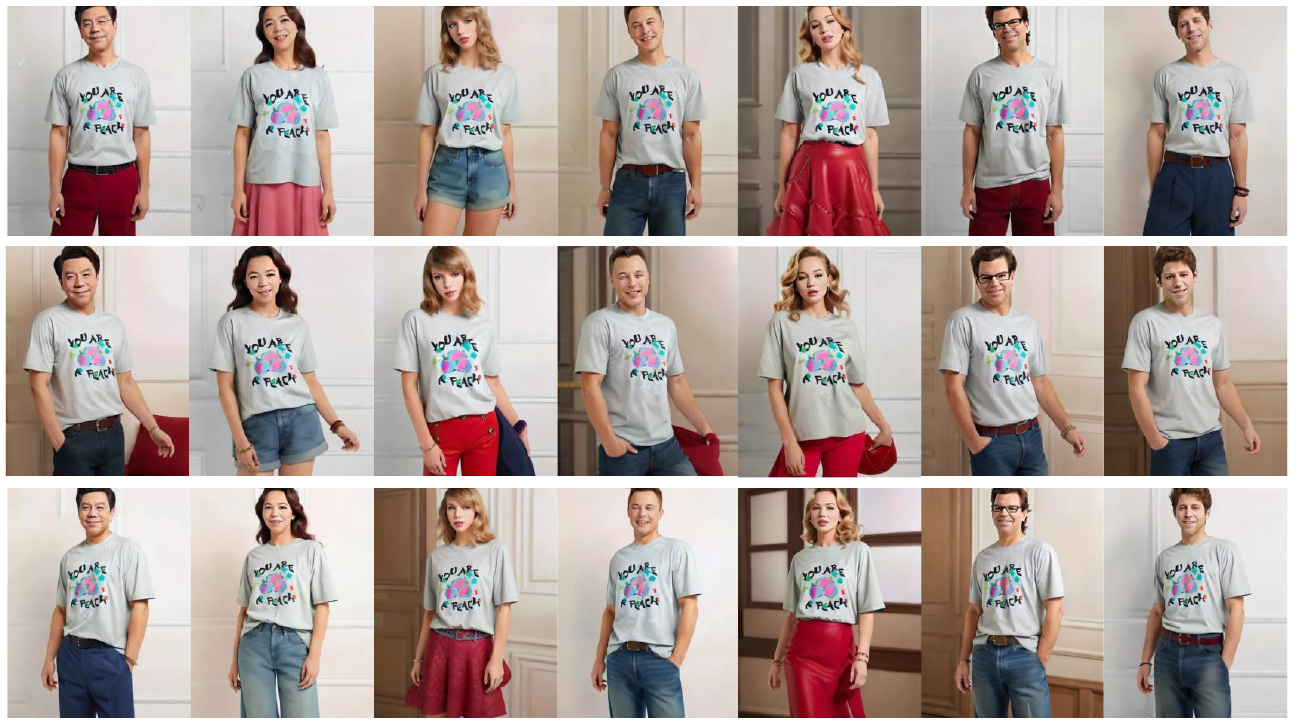}
\vspace{-0.3cm}
    \caption{More results of IMAGDressing-v1 synthesizing person images given garment with logos, and optional faces and pose.}
    \vspace{-0.3cm}
    \label{fig:more_face_pose}
\end{figure*}

\begin{figure*}[t]
    \centering
\includegraphics[width=\textwidth]{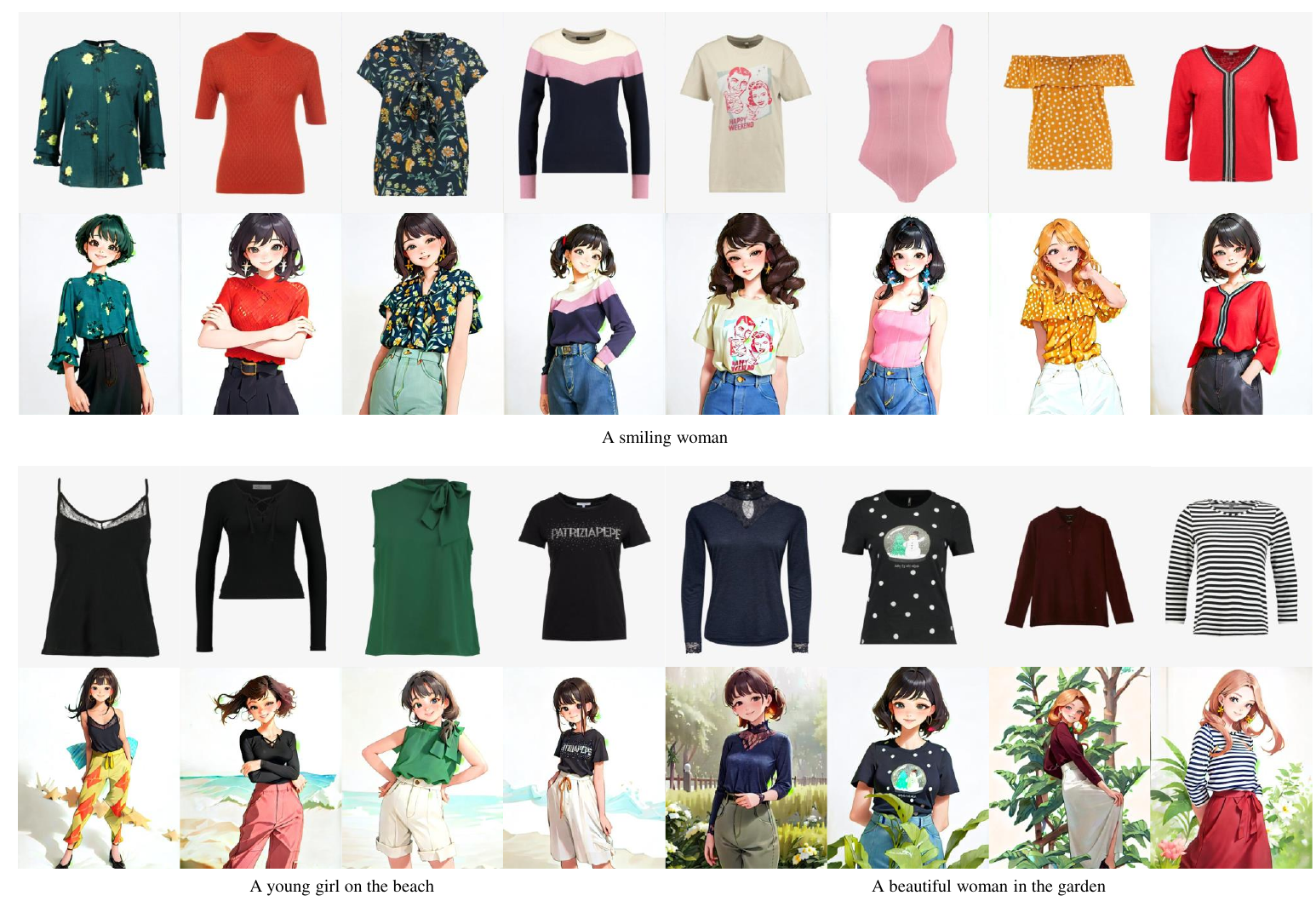}
\vspace{-0.5cm}
    \caption{More results of IMAGDressing-v1 synthesizing cartoon images based on clothing and different text prompts.}
    \vspace{-0.3cm}
    \label{fig:cartoon_style}
\end{figure*}

\end{document}